# A Type–II Fuzzy Entropy Based Multi-Level Image Thresholding Using Adaptive Plant Propagation Algorithm

Sayan Nag[*]

Department of Electrical Engineering,
Jadavpur University
Kolkata
India

[*]Corresponding Author

*Abstract-* **One of the most straightforward, direct and efficient approaches to Image Segmentation is Image Thresholding. Multi-level Image Thresholding is an essential viewpoint in many image processing and Pattern Recognition based real-time applications which can effectively and efficiently classify the pixels into various groups denoting multiple regions in an Image. Thresholding based Image Segmentation using fuzzy entropy combined with intelligent optimization approaches are commonly used direct methods to properly identify the thresholds so that they can be used to segment an Image accurately. In this paper a novel approach for multi-level image thresholding is proposed using Type II Fuzzy sets combined with Adaptive Plant Propagation Algorithm (APPA). Obtaining the optimal thresholds for an image by maximizing the entropy is extremely tedious and time consuming with increase in the number of thresholds. Hence, Adaptive Plant Propagation Algorithm (APPA), a memetic algorithm based on plant intelligence, is used for fast and efficient selection of optimal thresholds. This fact is reasonably justified by comparing the accuracy of the outcomes and computational time consumed by other modern state-of-the-art algorithms such as Particle Swarm Optimization (PSO), Gravitational Search Algorithm (GSA) and Genetic Algorithm (GA).**

*Keywords – Multi-level Image Segmentation, Image Thresholding, Fuzzy Entropy, Adaptive Plant Propagation Algorithm*

___________________________________

[*]nagsayan112358@gmail.com

# 1. Introduction

Image Segmentation is the process to discriminate between the objects and the background information. Various Image Segmentation strategies exist out of which Image Thresholding is a standout amongst the least complex and direct techniques which can effectively and efficiently segment an image to differentiate between the object and the background through a set of thresholds at pixel levels. This forms a part of the so-called approach of Intensity Based Segmentation. The automatic demarcation between objects and the background information remains the most challenging yet interesting domain of research for image processing and pattern recognition related works [1, 2]. Albeit different methodologies exist for Image Segmentation, Thresholding Segmentation is the most generally utilized proficient and least complex approach, appreciated by the researchers.

Bilevel Thresholding is a Two-level Thresholding technique where essentially two thresholds are selected and are used to separate the object from the background. Multilevel Thresholding is a method which can classify the pixels into various groups thus creating multiple regions eventually leading to finer segmentation results. In effect, an image is segmented into different objects in Multilevel Thresholding which significantly differs from the Bilevel Thresholding Approach and thus the former is sometimes preferred more than the latter [3]. Thresholding segmentation is important in image analysis and image understanding. Therefore it has wide applications in many areas like medical image analysis, image classification, object identification, feature extraction, image registration and pattern recognition, and so on.

The existing thresholding methods can be classified into two categories, namely, parametric and nonparametric [4, 5]. The main drawback of the parametric methods are excessive time consumption and the computational burden associated with these methods. On the other hand, the nonparametric approaches enjoys certain advantages. They are generally more robust and precise as compared to the parametric methods. As a consequence, these nonparametric approaches get more attention and are used widely and efficiently, as they can well determine the optimal set of thresholds by optimizing some existing standards like between-class variance, the entropy and the error rate which essentially contributes to the robustness and efficiency of these nonparametric methods.

A nonparametric multilevel image segmentation algorithm has been developed by Otsu [6]. Many entropy based algorithm has also been proposed in the recent years including the Shannon entropy [7], Tsalli's entropy [8], Renyi's entropy [9] and a fuzzy entropy based algorithm [10] which effectively contributed to the advancement in the field of Image Segmentation. But the demerits, like longer computational time and complexity, could not be avoided. Zhao et al. [11] proposed a state-of-the-art technique by thresholding the image using histogram partitions with certain fuzzy membership values and derived a criterion for selection of the optimal threshold. Tao et al. [12] proposed a fuzzy entropy based technique as a modification to the previously mentioned work. They used fuzzy partition for dividing the image histogram into various objects. A generalization of Type I fuzzy sets is Type II Fuzzy sets. A new measure called ultra-fuzziness is introduced in association with the image segmentation approach using Type II Fuzzy Sets. This new measure is important in the respect that it has been used to obtain the optimal image thresholds for the multi-level image thresholding operation [13, 14]. Algorithms for bi-level image segmentation as well as for multi-level segmentation using Type II Fuzzy Sets have been proposed in the earlier works, yet a lot remained to be explored. In this paper, a Type II Fuzzy based multi-level image segmentation algorithm is described. In order to obtain the best thresholds, a measure Type II Fuzzy entropy is maximized. With increase in the number of thresholds, the execution time will increase exponentially leading to computation burden. For this reason a novel modified Adaptive Plant Propagation Algorithm (APPA) is introduced which is used for fast convergence and less computational time. There exists a handful of other meta-heuristic like Genetic Algorithm (GA) [15], Particle Swarm Optimization (PSO) [16], Gravitational Search Algorithm (GSA) [17], Ant Colony Optimization (ACO) [18, 19], Stimulated Annealing (SA) [20, 21], Bacteria Forging Optimization (BFO), and so on [22, 23], but APPA stands out to be the best for optimizing the proposed fitness function, which is discussed later in Section 4.

The rest of the paper is organized as follows: A precise introduction to Type II Fuzzy Sets and the proposed algorithm of multi-level image segmentation is described in Section 2. Section 3 provides a brief introduction to Adaptive Plant Propagation Algorithm. Section 4 depicts the experimental results and comparisons. Finally, the conclusion is provided in Section 5.

## 2. Multilevel Thresholding Using Fuzzy Type II Sets

A Type-I fuzzy set A, in a finite set, $X = \{x_1, x_2, \ldots, x_n\}$ may be represented as

$$A = \{x, \mu_A(x) \mid x \in X, 0 \leq \mu_A(x) \leq 1\}$$

In Type-II fuzzy a range of membership values are used instead of a single value so that it can handle much uncertainty. It may be defined as:

$$A = \{x, \mu_A^{high}(x), \mu_A^{low}(x) \mid x \in X, 0 \leq \mu_A^{high}(x), \mu_A^{low}(x) \leq 1\} \quad (1)$$

where $\mu_A^{high}(x)$ and $\mu_A^{low}(x)$ are respectively the upper and lower membership functions.

Let a grayscale image $I$ be of the size of $M \times N$. Here $M$ represents the number of rows and $N$ is the number of columns. Let $f(m, n)$ represents the gray value of the pixel $(m, n)$ where $m \in \{1,2, \ldots, M\}$ and $n \in \{1,2, \ldots, N\}$. If $L$ is the number of gray levels by which image $I$ can be represented then the set of all gray levels $\{0, 1, 2 \ldots, L-1\}$ is represented as $GL$:

$$f(m, n) \in GL \text{ for all } (m, n) \in I$$
$$D_k = \{(m,n) : f(m,n) = k, \text{ where } k \in \{0, 1, 2 \ldots, L-1\}\} \quad (2)$$

Let us consider $H = \{h_0, h_1, \ldots, h_{L-1}\}$ to be the normalized histogram of the image $I$ where $h_k = n_k / (M \times N)$, where $n_k$ is the number of pixels present in $D_k$. A measure ultra-fuzziness is introduced in association with the fuzzy set. It gives a 0 value when the membership values can be defined without any uncertainty. While on the other hand the value increases to 1 when membership values can be indicated within an interval. The ultra-fuzziness for the kth level can be defined mathematically as follows:

$$P_k = \sum_{i=0}^{L-1} \left(h_i * (\mu_k^{high}(i) - \mu_k^{low}(i))\right)$$

Here $k \in \{1,2, \ldots, LV + 1\}$, $LV$ is the number of levels of segmentation and $\mu_k$ is the trapezoidal fuzzy membership function for gray values to belong to $k^{th}$ level out of $LV + 1$ levels of segmentation. It is defined as follows:

$$\mu_k(i) = \begin{cases} 0, & i \leq a_{k-1} \\ \dfrac{i - a_{k-1}}{c_{k-1} - a_{k-1}}, & a_{k-1} < i \leq c_{k-1} \\ 1, & c_{k-1} < i \leq a_k \\ \dfrac{i - C_k}{a_k - C_k}, & a_k < i \leq c_k \\ 0, & i > c_k \end{cases}$$

Here $a_k$ and $c_k$, $k \in \{1,2, \ldots, LV + 1\}$, are the fuzzy parameters. It is to be noted that $a_0 = c_k = 0$ and $a_{LV+1} = c_{LV+1} = L - 1$. $LV$ denotes the number of thresholds as represented before. The fuzzy type-II entropy for kth level of segmentation is defined as follows:

$$H_k = -\sum_{i=0}^{L-1} \left(\dfrac{\left(h_i * \left(\mu_k^{high}(i) - \mu_k^{low}(i)\right)\right)}{P_k}\right) * \log_e \left(\dfrac{\left(h_i * \left(\mu_k^{high}(i) - \mu_k^{low}(i)\right)\right)}{P_k}\right) \quad (3)$$

where $k \in \{1,2, \ldots, LV + 1\}$, the net entropy is given as the summation of entropies of all the levels.

$$H(a_1, c_1, \ldots, a_n, c_n) = \sum_{k=1}^{N+1} H_k$$

In order to obtain the optimal fuzzy parameters the total entropy needs to be maximized. So $(a_1, c_1, \ldots\ldots, a_n, c_n)^*$ be that optimal zet of fuzzy parameters which maximizes *H*. Thus the threshold for the image *I* is given by the following equation:

$$T_n = 0.5 * (a_n + c_n) \tag{4}$$

where $n = \{1, 2, \ldots\ldots, LV\}$, *LV* denotes the number of thresholds. The parameter set $(a_1, c_1, \ldots\ldots, a_n, c_n)$ acts as an individual in the proposed novel modified Adaptive Plant Propagation Algorithm which is described in the following section.

## 3. Adaptive Plant Propagation Algorithm (APPA)

The Multi-objective Optimization gained popularity essentially because in the real world applications we deal with such problems where we have to pacify a number of objectives along with certain curtailments and they should be satisfied simultaneously for optimum performance. Traditional methods include Gradient Descent, Dynamic Programming and Newton Methods which are used earlier to solve these Multi-objective Optimization problems. But the major drawback of these traditional approaches is that they are computationally less efficient. Keeping in mind the complexity of the non-linear optimization problems, the meta-heuristic algorithms such as GA, PSO, ABC, and ACO came into existence. These meta-heuristic approximate algorithms provide better solutions, at the same time being computationally efficient, to these kind of optimization problems. Such an approximate optimization algorithm is Plant Propagation Algorithm (PPA) also known as Strawberry Algorithm.

Strawberry Algorithm is a nature inspired new meta-heuristic algorithm which imitates the propagation of the Strawberry Plant. Scientific studies evince the intelligent and exciting behaviors exhibited by the plants. Plants have an innate tendency to interact with the external world with inherent defense mechanisms in order to counteract the exploitation by insects like caterpillars. Root grows deeper, under the ground, the light and nutrient information reach the growth center in root tips and the root is oriented accordingly based on the directions. Hence, plants are really intelligent and thus attracts researchers to make use of their growth and defense mechanisms in order to develop algorithms based on their intelligence. Various algorithms like Flower Pollination Algorithm (FPA) [24 - 26] and Root Shoot Growth Coordination Optimization (RSCO) exist based on the properties revealed by the plants.

Unlike Simulated Annealing (SA), the Strawberry Algorithm also known as the Plant Propagation Algorithm (PPA) is a multipath following algorithm [27 - 30]. Effectively, PPA is not restricted to a single path, rather there exist a number of multiple paths in which the solution may be directed. Thus, it is bound to have better exploitation and exploration properties which is not present in the Simulated Annealing (SA). Exploitation is defined as that property by which an algorithm makes a search closer to optimum solutions whereas exploration indicates the covering of the search region. The chances of getting a good solution exalt with the betterment of these two properties.

Plants propagate through runners. Plants optimize its survival based on certain basic necessities including the availability of water, nutrients, light and toxic substances. In the event that a plant is available in a place where its underlying foundations are in a decent area under the ground with accessibility of enough supplements and water then it will scarcely depart the position for its survival as long as the convergence of water and minerals in that spot stays pretty much according to the necessity. So there will be an inclination to send short runners giving new strawberry plants, in this way, involving the area as much as it can. Presently, for another illustration, if a plant is in a place without the fundamental necessities like light, water and supplements then it will attempt to locate a superior spot for its posterity by sending few long runners encourage at longer separations for misusing the removed neighborhoods to locate an ideal place for its survival. Additionally the quantity of long runners are a couple since it winds up plainly costly for the plant to send long runners particularly when it is in a poor area without any supplements and water. Natural elements chooses a place, regardless of whether it is great or awful, whether it has ample supplements and water fixations or not. These elements thus ponder the development of plants and their supportability. This fundamental proliferation technique created after some time for guaranteeing survival in species gets consideration and this is the means by which Strawberry Algorithm or when all is said in done, Plant Propagation Algorithm is developed which mirrors the engendering of plants. A set of parameters and function is required: population estimate, a fitness function, the quantity of runners to be made for every arrangement and the range of allowable distance for every runner.

A plant is considered to be in a location $Y_i$ where the dimension of the search space is given as $n$. So, essentially $Y_i = \{y_{i,j}, \text{ for all } j = 1,\ldots,n\}$. Let the population size be denoted as $N_p$ which determines the number of strawberry plants to be used initially. It is known that Strawberry plants which are in poor spots propagate by sending long runners which are few in number, the process being known as exploration and the plants which are in location with abundance of essential nutrients, minerals and water propagate by sending many short runners, the process beingis known as exploitation). Maximum number of generations considered is $g_{max}$ (stopping criterion in the algorithm) and maximum number of permissible runners per plant is $n_{max}$.

The objective function values at different positions $Y_i$, $i = 1,\ldots,N_p$ are calculated. These possible candidate solutions will be sorted according to their fitness scores. Here the fitness is a function of value of the objective function under consideration. It is better to keep the fitness scores within a certain boundary between 0 and 1, that is, if the fitness function is $f(x) \in [0, 1]$. To keep the fitness values within this range a mapping is done which involves another function, known as the *sigmoid function*, described as:

$$N(x) = \frac{\exp\left(\frac{f(x)}{\max(f(x))}\right)}{1+\left(\exp\left(\frac{f(x)}{\max(f(x))}\right)\right)} \quad (5)$$

The number of runners that are determined by the functions and the distance of propagation of each of them are described. There exists a direct relation between the number of runners produced by a candidate solution and its fitness given as:

$$n_r = ceil(n_{max} \; N_i \; r) \quad (6)$$

Here, $n_r$ is the number of runners produced for solution $i$ in a particular generation or iteration after the population is sorted according to the fitness given in (5), $n_{max}$ is the number of runners which is maximum permissible, $N_i$ is the fitness as shown in (5), $r$ is just a random number lying between 0 and 1 which is randomly selected for each individual in every iteration or generation and *ceil*() refers to the ceiling function. The minimum number of runners is 1 and maximum is $n_r$. This function safely ensures that at least 1 runner should be there which may correspond to the long runner as described before. The length of each runner is inversely related to its growth as shown below:

$$d_j^i = (1 - N_i)(r - 0.5), \text{ for } j = 1,\ldots, n \quad (7)$$

where, $n$ represents the dimension of the search space. So, each runner is restricted to a certain range between -0.5 and 0.5. The calculated length of the runners are used to update the solution for further exploration and exploitation of the search space by the equation:

$$x_{i,j} = y_{i,j} + (b_j - a_j) d_j^i, \text{ for } j = 1,\ldots, n \quad (8)$$

The algorithm is modified to be an adaptive one in view of the limits of the search domain. Hence, the name is given as Adaptive Plant Propagation Algorithm (APPA) or Strawberry Algorithm (ASA). In the event that the limits are disregarded the point is changed in accordance to lie within the search space. Essentially, $a_j$ and $b_j$ are the respective lower and upper boundaries of the $j^{th}$ coordinate of the search space. After every single individual plant in the population have passed on their designated runners, new plants are surveyed and the whole extended population is arranged. To keep the population fixed, rather the size of the population fixed, it is to be guaranteed that the candidates with lower growth are dispensed from the population. Another strategy is adopted to to abstain from being struck in the local minima. It might happen that for a certain number of back to back generations there is no improvement in a candidate solution, rather the runners it sends out are also not fit to remain in the population. So a threshold to be set for such a solution such that if the number of generations in which it is not enhancing surpasses the threshold then the solution is discarded and another fresh candidate solution or individual is produced within the limits of the pursuit space.

In our problem, this proposed Adaptive Plant Propagation Algorithm (APPA) is applied with constraints as and cost function. All the equality and inequality constraints are taken care of; the inequality constraints basically defines the search space of the approximate solution. The following picture is a representation of the algorithm with all the steps to be performed in detail.

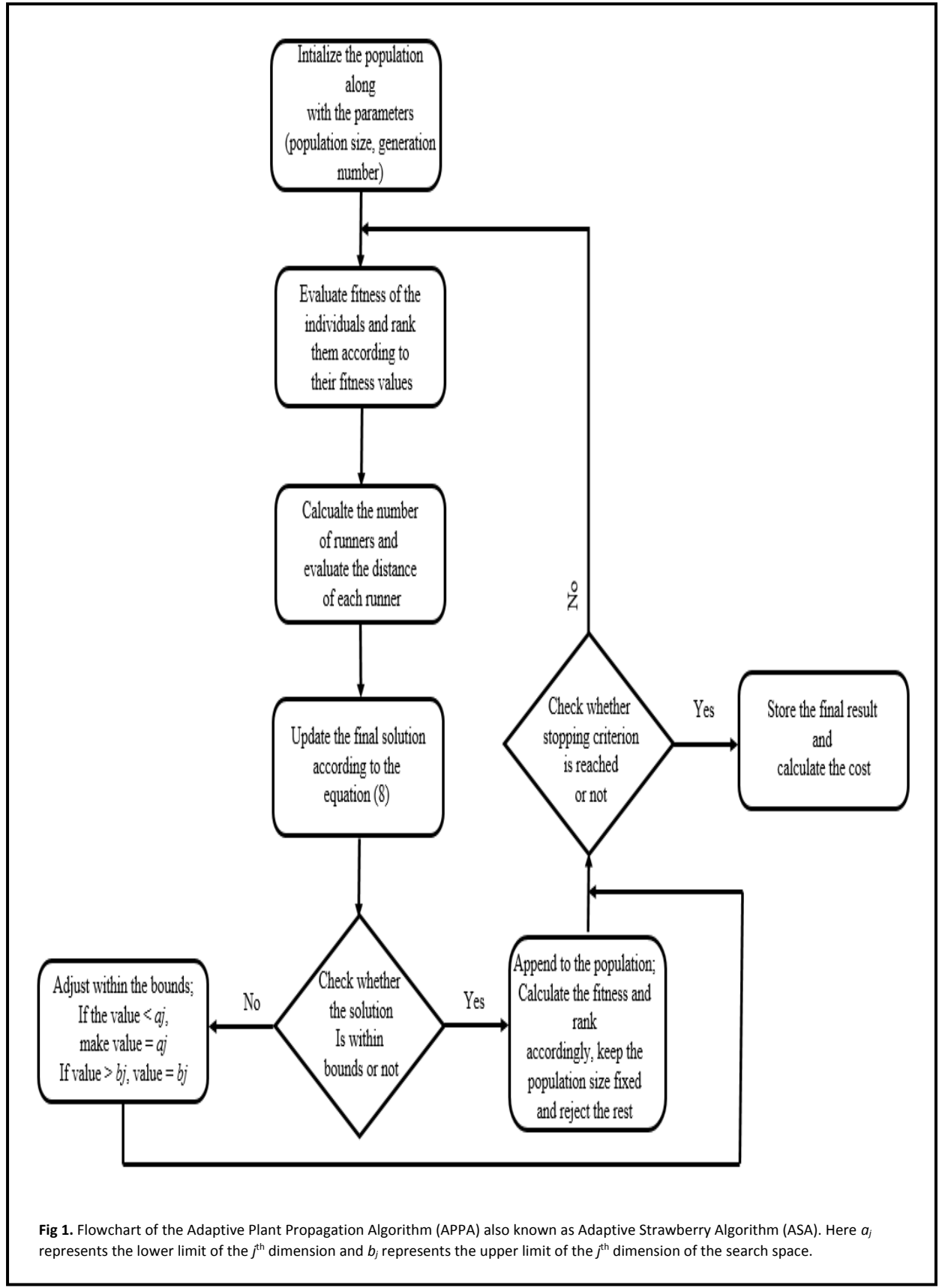

**Fig 1.** Flowchart of the Adaptive Plant Propagation Algorithm (APPA) also known as Adaptive Strawberry Algorithm (ASA). Here $a_j$ represents the lower limit of the $j^{th}$ dimension and $b_j$ represents the upper limit of the $j^{th}$ dimension of the search space.

## 4. Experimental Results

The simulations are performed in MATLAB 2013a in a workstation with Intel Core i3 2.9 GHz processor. In order to test and compare the efficiency and robustness of the proposed approach to multi-level image segmentation, four images have been chosen arbitrarily from Berkeley Segmentation Dataset and Benchmark. The Table for the setup of the APPA Algorithm is given below.

**Table 1.** APPA Set-up

| Parameters | Values |
|---|---|
| Number of Runs | 10 |
| No. of iterations per run | 100 |
| Dimension of the search space($D$) | 2*No. of Thresholds |
| Upper bound of search space | Maximum gray level present in the image |
| Lower bound of search space | Minimum gray level present in the image |
| Number of Particles($N_p$) | 10*$D$ |
| Maximum Number of Roots considered($n_{max}$) | 3 |

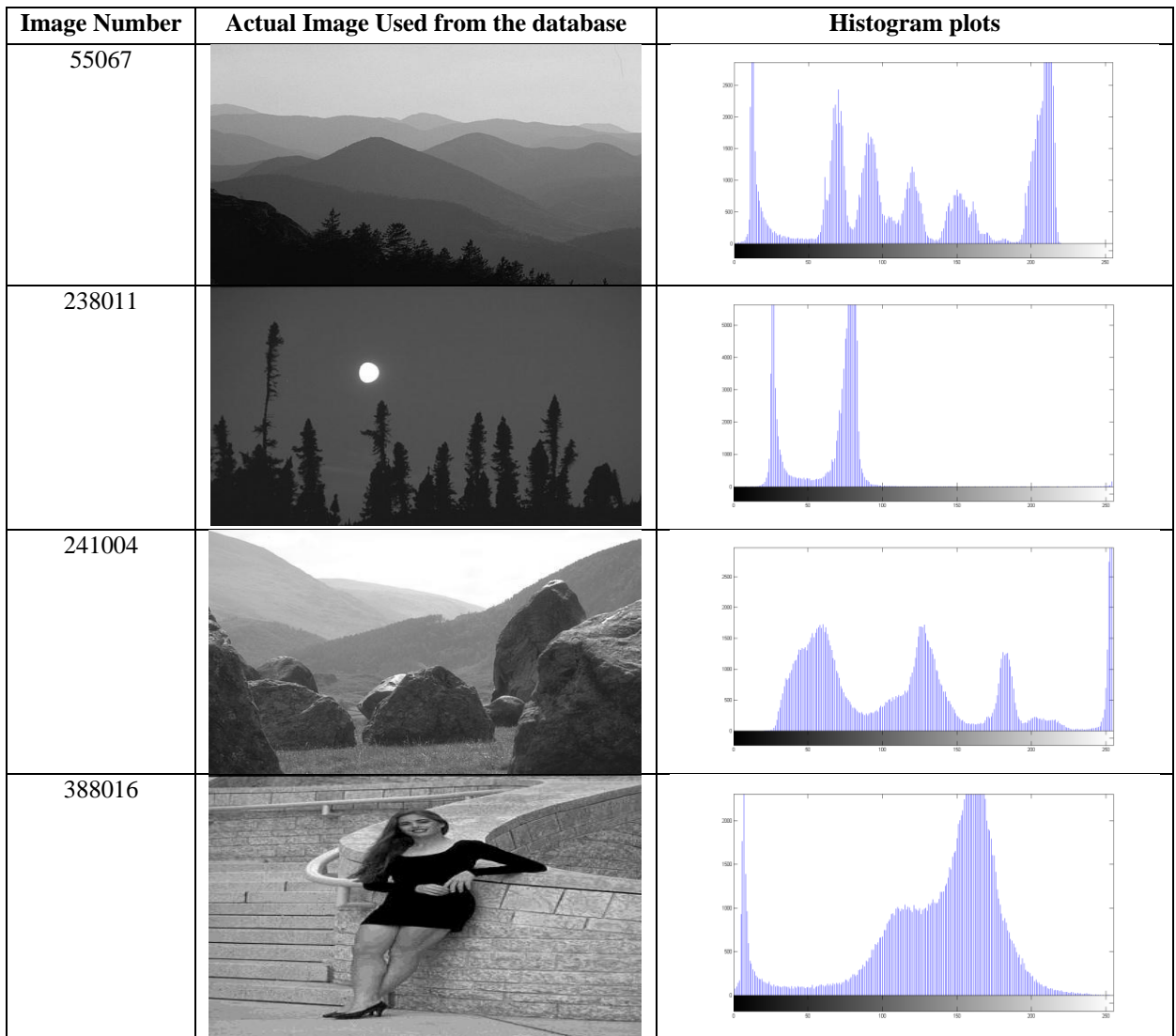

**Fig 2.** Test Images from Berkeley Segmentation Dataset and their Histogram Plots

We have shown the test images thresholded by the proposed algorithm below for 2, 3, 4 thresholds. The image number associated with each image corresponds to the number provided in the Berkeley Segmentation Dataset. The fuzzy membership function parameters and the corresponding thresholds for the test images are listed in table 2 for the proposed method.

**Table 2.** Fuzzy parameters and thresholds for images thresholded by Fuzzy Type II method

| Image No. | No. of thresholds ($T_N$) | Fuzzy Parameters | Thresholds |
|---|---|---|---|
| 55067 | 2 | 13, 61, 66, 219 | 37, 143 |
|  | 3 | 12, 68, 71, 89, 90, 212 | 40, 80, 151 |
|  | 4 | 12, 60, 60, 93, 97, 130, 145, 203 | 36, 77, 114, 174 |
| 238011 | 2 | 27, 73, 101, 151 | 50, 76 |
|  | 3 | 22, 90, 98, 98, 105, 251 | 56, 98, 178 |
|  | 4 | 32, 75, 90, 90, 107, 151, 153, 254 | 54, 90, 129, 204 |
| 241004 | 2 | 22, 185, 185, 252 | 104, 219 |
|  | 3 | 28, 148, 150, 174, 179, 253 | 88, 162, 216 |
|  | 4 | 25, 61, 63, 146, 152, 169, 169, 252 | 43, 105, 161, 211 |
| 388016 | 2 | 2, 158, 163, 251 | 80, 207 |
|  | 3 | 4, 100, 108, 154, 156, 251 | 52, 131, 204 |
|  | 4 | 8, 86, 89, 90, 96, 148, 149, 235 | 47, 90, 122, 192 |

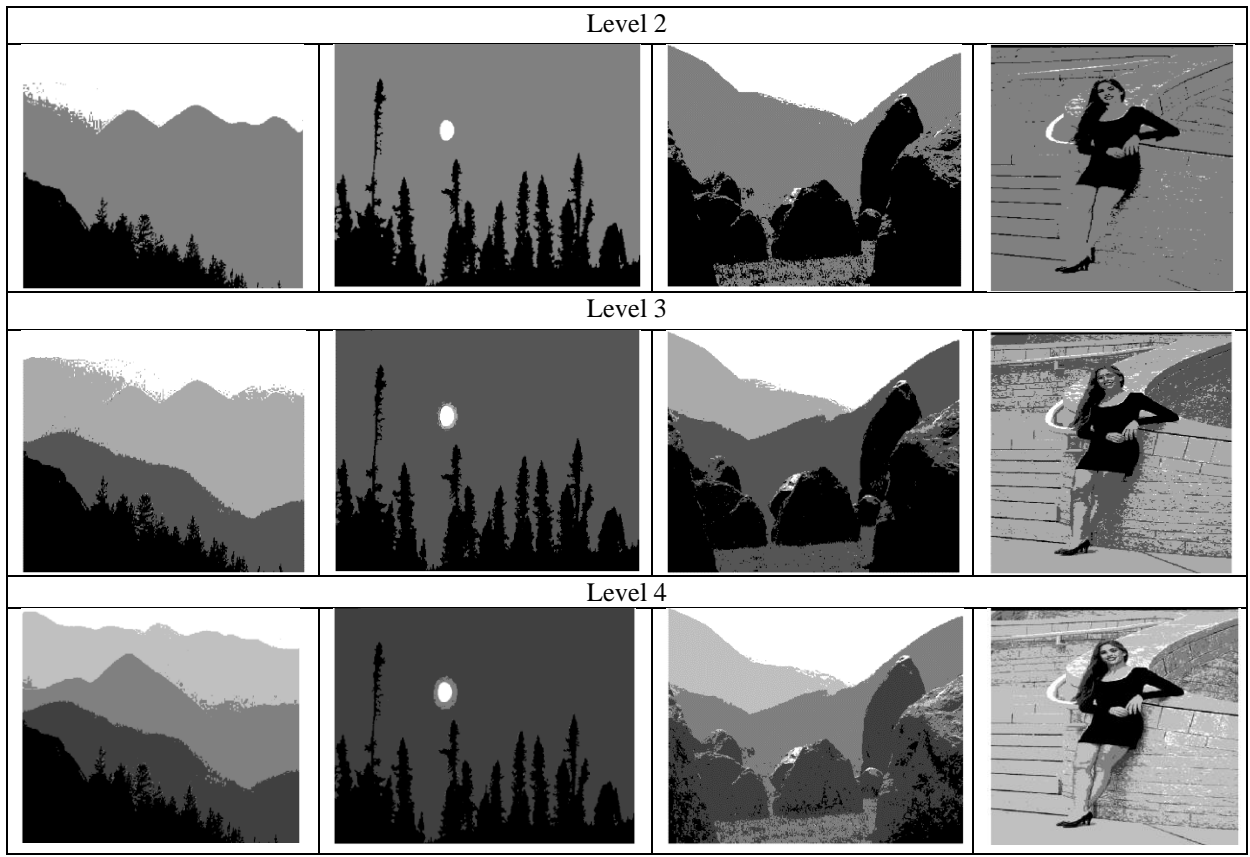

**Fig 3.** Test Images from Berkeley Segmentation Dataset segmented by Fuzzy Type-II combined with APPA (proposed method)

The images shown is in accordance with the order mentioned in the table (order is 55067, 238011, 241004 and then 388016). It is observed that the bushes merged with the mountains are properly segmented by the proposed algorithm in the figure corresponding to image number 55067, level 3. Also the transitions in mountains are more or less clearly defined due to proper choice of thresholds done by the proposed algorithm which perfectly segments the sky, mountains and the bushes. In the figure corresponding to image number 238011, level 3, the moon and the trees and the sky are properly segmented by the proposed method. In the figure corresponding to image number 241004, level 3, the mountain properly demarcated from the sky and also the rocks are accurately segmented from the surroundings by the proposed Fuzzy Type II method. It is to be noted from the original image corresponding to image number 241004, that there exist two mountain ranges with different gray levels; these are properly segmented from each other with increase in the number of thresholds that is done by the proposed algorithm. In the figure corresponding to image number 388016, level 3, the lady is properly segmented from the background consisting of walls, the task being performed by the proposed algorithm. A close look into the histograms of each of these images will reasonably justify the accuracy of the optimized thresholds. The thresholds in most of the cases lie in the saddle points of the histograms of the images considered and the accuracy of determination of these thresholds increases with increase in the number of levels although one thing should be taken care of that the number of levels should also correspond to the image histogram which can be judged by eye estimation.

**Table 3.** Mean fitness function value and its standard deviation

| Image No. | $T_N$ | GSA | | PSO | | GA | | APPA | |
|---|---|---|---|---|---|---|---|---|---|
| | | $f_{mean}$ | Std. | $f_{mean}$ | Std. | $f_{mean}$ | Std. | $f_{mean}$ | Std. |
| 55067 | 2 | 13.0160 | 8.47E-02 | 13.2502 | 1.12E-01 | 13.0923 | 1.03E-01 | **13.2735** | **1.99E-10** |
| | 3 | 16.1281 | 1.61E-01 | 16.4828 | 1.25E-01 | 16.5737 | 1.29E-01 | **16.6988** | **5.42E-07** |
| | 4 | 19.2494 | 1.87E-01 | 19.2815 | 2.03E-01 | 19.1033 | 1.20E-01 | **19.2996** | **2.57E-10** |
| 238011 | 2 | 11.3416 | 1.22E-01 | 11.0577 | 2.26E-02 | 11.3668 | 2.47E-01 | **11.8458** | **2.41E-08** |
| | 3 | 14.8768 | 1.08E-01 | 15.0379 | 2.32E-01 | 15.0743 | 1.55E-01 | **15.4475** | **8.31E-14** |
| | 4 | 17.8144 | 2.89E-01 | 17.8814 | 1.24E-01 | 18.1718 | 1.81E-01 | **18.2837** | **2.95E-07** |
| 241004 | 2 | 13.5155 | 4.59E-02 | 13.5452 | 2.18E-02 | 13.5538 | 7.17E-02 | **13.6432** | **6.89E-06** |
| | 3 | 17.2157 | 2.66E-01 | 17.2599 | 1.22E-01 | 17.1973 | 2.09E-01 | **17.4898** | **3.75E-15** |
| | 4 | 20.3838 | 1.48E-01 | 20.4251 | 1.32E-01 | 20.4055 | 4.38E-02 | **20.4468** | **1.19E-13** |
| 388016 | 2 | 13.0098 | 7.32E-02 | 13.1623 | 3.22E-02 | 13.2421 | 9.29E-02 | **13.6917** | **8.23E-11** |
| | 3 | 16.9898 | 8.61E-02 | 17.0012 | 1.55E-01 | 17.1331 | 5.75E-02 | **17.3583** | **8.35E-14** |
| | 4 | 20.6309 | 8.91E-02 | 20.8155 | 1.08E-01 | 20.7835 | 6.44E-02 | **21.0333** | **3.12E-09** |

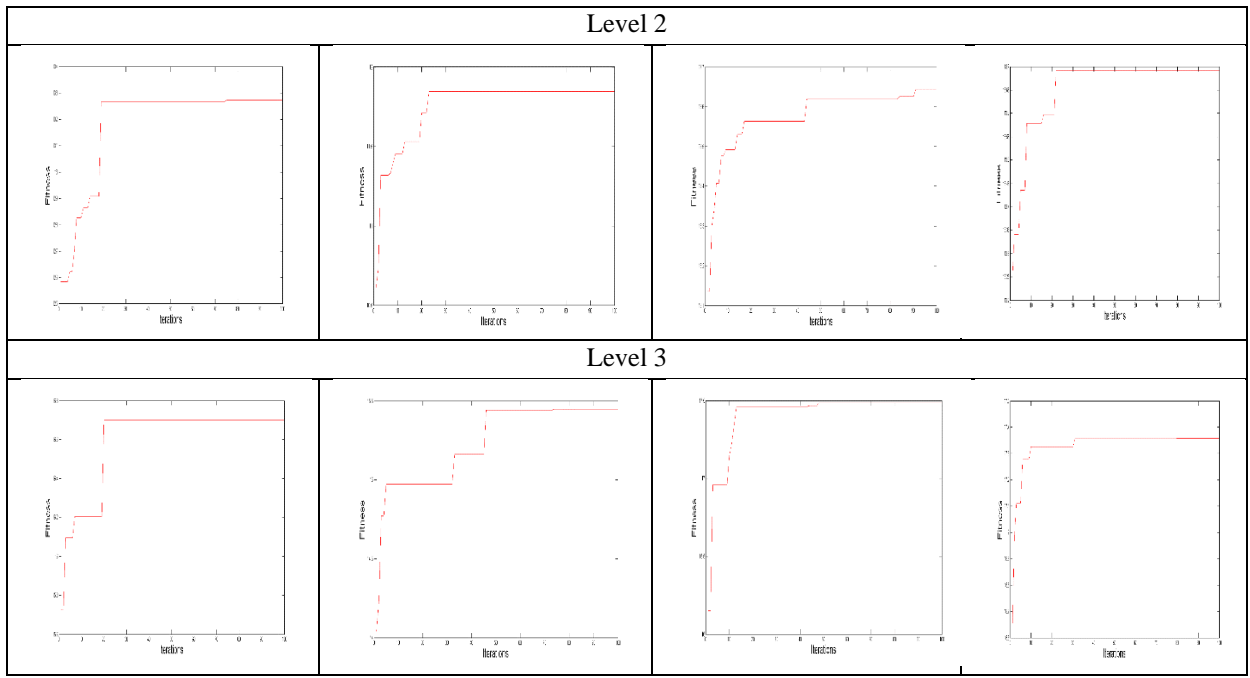

| Level 4 |
|---|

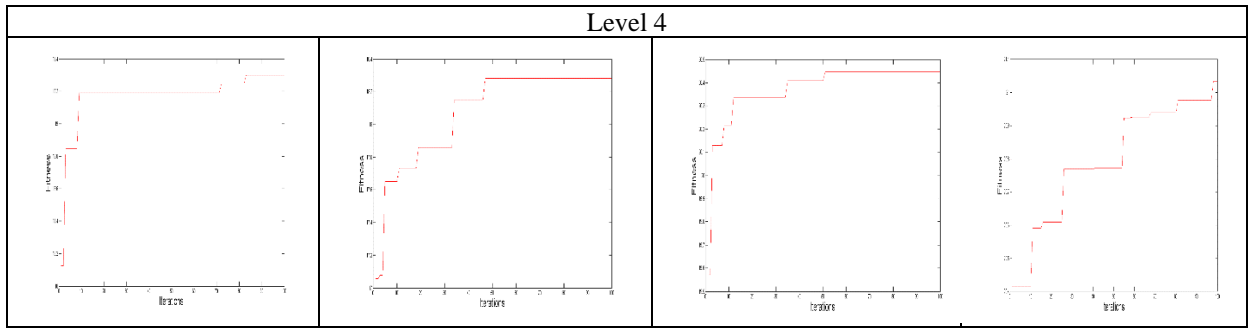

**Fig 4.** Convergence plots for Images 55067, 238011, 241001 and 388016 (in this order from left to right horizontally) for 2, 3 and 4 levels

Performance of APPA is juxtaposed with some advanced algorithms like Particle Swarm optimization (PSO), Gravitational Search Algorithm (GSA) and Genetic Algorithm (GA). We have only considered the rudimentary algorithms of all the optimizers and not their variations. In the table 3 the mean fitness values along with the standard deviation values evaluated over 10 runs each comprising of 100 iterations are listed. It is observed from the experimental results that the optimum fitness values for all the images are maxima for APPA with respect to the other optimizers; this is a significant achievement for the proposed approach. It may also be noticed that the standard deviations corresponding to the optimum fitness values are minima for all the images when optimized by APPA. It is significant to mention that the average computation time consumed by APPA is lesser than that consumed by PSO, GSA and GA. In all cases APPA has maximum optimized fitness value or score with lesser standard deviations. The x-axis of the convergence plot indicates the number of iterations per run and the y-axis represents the fitness values achieved at every iteration. The convergence plots for the test images for levels 2, 3 and 4 also reveals the efficiency and robustness of the proposed method used for multi-level segmentation of images.

## 5. Conclusion and Future Works

The proposed algorithm of image thresholding by Type II Fuzzy Sets accompanied with APPA undoubtedly segments the image into different objects accurately and in the meantime devours as less time as conceivable to fulfill the task. The proposed algorithm based on Type II Fuzzy Entropy produces better results with APPA than other existing techniques like GSA, GA and PSO. The use of APPA not only speeds up the search process but also significantly contributes to the accuracy of the outcomes. Thus we can say that APPA stands out to be one such dominating plant-intelligence meta-heuristics which is able to obtain better image segmentation quality and also showing obvious superiority in terms of running time and convergence speed. Future works include improvisation of this algorithm to segment the images more precisely devouring lesser time which is of prime concern for fast real time applications involving image segmentation tasks.